\documentclass[10pt,twocolumn,letterpaper]{article}

\usepackage{cvpr}
\usepackage{times}
\usepackage{epsfig}
\usepackage{graphicx}
\usepackage{amsmath}
\usepackage{amssymb}
\usepackage{enumitem}
\usepackage{color}

\usepackage[pagebackref=true,breaklinks=true,letterpaper=true,colorlinks,bookmarks=false]{hyperref}

\cvprfinalcopy 

\def\cvprPaperID{2312} 

\ifcvprfinal\fi
\begin{document}

\title{Learning a Discriminative Filter Bank\\within a CNN for Fine-grained Recognition}

\author{Yaming Wang$^{1}$, \; Vlad I. Morariu$^{2*}$, \; Larry S. Davis$^1$\\
$^1$University of Maryland, College Park \quad $^2$Adobe Research\\
\quad\quad\quad {\tt\small \{wym, lsd\}@umiacs.umd.edu} \quad\quad\quad {\tt\small morariu@adobe.com}
}

\maketitle
\thispagestyle{empty}
\let\thefootnote\relax\footnote{*The work was done while the author was at University of Maryland.}
\begin{abstract}
Compared to earlier multistage frameworks using CNN features, recent end-to-end deep approaches for
fine-grained recognition essentially enhance the mid-level learning capability of CNNs. Previous approaches achieve
this by introducing an auxiliary network to infuse localization information into the main classification network, or
a sophisticated feature encoding method to capture higher order feature statistics. We show that mid-level
representation learning can be enhanced within the CNN framework, by learning a bank of convolutional
filters that capture class-specific discriminative patches without extra part or bounding box annotations. Such a filter
bank is well structured, properly initialized and discriminatively learned through a novel asymmetric multi-stream
architecture with convolutional filter supervision and a non-random layer initialization. 
Experimental results show that our approach achieves state-of-the-art on three publicly available fine-grained recognition datasets
(CUB-200-2011, Stanford Cars and FGVC-Aircraft). Ablation studies and visualizations are provided
to understand our approach.
\end{abstract}

\section{Introduction} \label{sec1}
Fine-grained object recognition involves distinguishing sub-categories of the same super-category
(\eg, birds \cite{cub2011}, cars \cite{car196} and aircrafts \cite{fgvc_air}),
and solutions often utilize information from localized regions to capture subtle differences. Early
applications of deep learning to this task built traditional multistage frameworks upon convolutional neural network
(CNN) features; more recent CNN-based approaches are usually trained end-to-end and can be roughly divided into two
categories: \textit{localization-classification sub-networks} and
\textit{end-to-end feature encoding}.

Previous multistage frameworks utilize low-level CNN features to find discriminative regions or semantic parts,
and construct a mid-level representation out of them for classification \cite{krause15, 2attention, neural_act,
mgd, deepresp, tripmine}. These methods achieve better performance compared to two types of baselines: \textit{(i)}
they outperform their counterparts with hand-crafted features
(\eg, SIFT) by a huge margin, which means that low-level CNN features are far more effective than previous hand-crafted ones;
\textit{(ii)} they significantly outperform their baselines which finetune the same CNN used for feature extraction.
This further suggests that CNN's ability to learn mid-level representations is limited and still has sufficient room to improve.
Based on these observations, end-to-end frameworks aim to \textit{enhance the mid-level representation learning capability of CNN}.

The first category, \textit{localization-classification sub-networks}, consists of a
classification network assisted by a localization network. The mid-level learning capability of the
classification network is enhanced by the localization information (\eg part locations or segmentation masks) provided by 
the localization network. Earlier works from this category \cite{fg_rcnn, deeplac, spda_cnn, partstack, maskcnn} depend
on additional semantic part annotations, while more recent ones \cite{stn, taomei1, taomei2} only require category
labels. Regardless of annotations, the common motivation behind these approaches is to first \textit{find the corresponding parts} and then
\textit{compare their appearance}. The first step requires the semantic
parts (\eg head and body of birds) to be shared across object classes, encouraging
the representations of the parts to be similar; but, in order to be discriminative, the latter encourages the part representations to be
different across classes. This subtle conflict implies a 
trade-off between recognition and localization ability, which might reduce a single integrated network's
classification performance. Such a trade-off is also reflected in practice, in that training usually
involves alternating optimization of the two networks or separately training the two followed by joint tuning.
Alternating or multistage strategies complicate the tuning of the integrated network.
\begin{figure*}[tb]
\begin{center}
\includegraphics[width=0.8\textwidth]{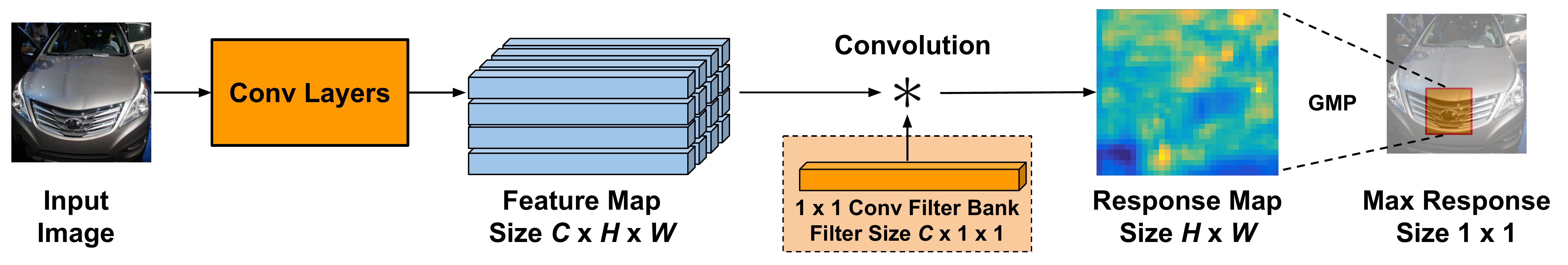}
\end{center}
\vspace{-9pt}
\caption{\label{fig1}The motivation of our approach is to regard a $C\times1\times1$ vector in a feature map as the representation of a small
patch and a $1\times1$ convolutional filter as a discriminative patch detector. A discriminative patch can be discovered
by convolving the feature map with the $1\times1$ filter and performing Global Max Pooling (GMP) over the response
map. The full architecture is illustrated in Figure \ref{fig2}.}
\end{figure*}

The second category, \textit{end-to-end feature encoding} \cite{b_cnn, c_b_cnn, lowrank_bcnn, highorder1, highorder2},
enhances CNN mid-level learning by encoding higher order statistics of convolutional feature maps. The need
for end-to-end modeling of higher order statistics became evident when the Fisher Vector encodings of
SIFT features outperformed a finetuned AlexNet by a large margin on fine-grained recognition \cite{fgvcfisher}. The resulting
architectures have become standard benchmarks in the literature. While effective, end-to-end encoding networks are
less human-interpretable and less consistent in their performance across non-rigid
and rigid visual domains, compared to localization-classification sub-networks.

This paper addresses the issues facing both categories of end-to-end networks. Our main contribution is to explicitly learn
discriminative mid-level patches \textit{within} a CNN framework in an end-to-end fashion without extra part or
bounding box annotations. This is achieved by regarding $1\times1$ filters as small ``patch detectors'', designing an asymmetric
multi-stream structure to utilize both patch-level information and global appearance, and introducing filter supervision with
non-random layer initialization to activate the filters on discriminative patches. Conceptually, our discriminative patches
differ from the parts in localization-recognition sub-networks, such that they are not necessarily shared across classes as long as they have
discriminative appearance. Therefore, our network fully focuses on classification and avoids the trade-off between recognition
and localization. Technically, a convolutional filter trained as
a discriminative patch detector will only yield a high response at a certain region for one class.

The resulting framework enhances the mid-level learning capability of the classical CNN by introducing a bank of
discriminative filters. In practice, our framework preserves the
advantages of both categories of previous approaches:
\begin{itemize}[noitemsep, topsep=0pt]
\item Simple and effective. The network is easy to build and once initialized only involves single-stage training. It
outperforms state-of-the-art.
\vspace{2pt}
\item High human interpretability. This is shown through various ablation studies and visualizations of learned
discriminative patches.
\vspace{2pt}
\item Consistent performance across different fine-grained visual domains and various network architectures.
\end{itemize}

\section{Related Work} \label{sec2}
\noindent\textbf{Fine-grained recognition}\quad Research in fine-grained recognition has shifted from multistage
frameworks based on hand-crafted features \cite{ningzhang3, berg1, bangpeng, symbiotic, fgvcfisher} to
multistage framework with CNN features \cite{krause15, 2attention, neural_act, deepresp, tripmine}, and then to
end-to-end approaches. 
Localization-classification sub-networks
\cite{fg_rcnn, deeplac, spda_cnn, partstack, stn, hsnet, taomei1} have a localization network which is usually a variant of R-CNN
\cite{rcnn, fast_rcnn}, FCN (Fully Convolutional Network) \cite{fcn} or STN (Spatial Transformer Network) \cite{stn} and
a recognition network that performs recognition based on localization.
More recent advances explicitly regress the location/scale of the parts using a recurrent localization network such as LSTM
\cite{hsnet} or a specifically designed recurrent architecture \cite{taomei1}. End-to-end encoding approaches
 \cite{b_cnn, c_b_cnn, lowrank_bcnn, highorder1, highorder2} encode higher order information. The classical benchmark, Bilinear-CNN
\cite{b_cnn} uses a symmetric two-stream
network architecture and a bilinear module that computes the outer product over the outputs of the two streams to capture
the second-order information. \cite{c_b_cnn} further observed that similar performance can be achieved by taking the
outer product over a single-stream output and itself.
More recent advances reduce high feature dimensionality \cite{c_b_cnn, lowrank_bcnn} or extract higher order
information with kernelized modules \cite{highorder1, highorder2}. Others have explored directions such
as utilizing hierarchical label structures \cite{fzhou1}, combining
visual and textual information \cite{fzhou2, embed_fg, lan_fgvc}, 3D-assisted recognition \cite{car196, yenliang,
boxcar}, introducing humans in the loop \cite{human1, human2, human3}, and collecting larger amount of data \cite{har,
large_car, unreason, vegfru}.

\noindent\textbf{Intermediate representations in CNN}\quad Layer visualization \cite{fergus14} has shown that the
intermediate layers of a CNN
learn human-interpretable patterns from edges and corners to parts and objects. Regarding the discriminativeness of such
patterns, there are two hypotheses. The first is that some neurons in these layers behave as ``grandmother cells''
which only fire at certain categories, and the second is that the neurons forms a distributed code where the firing
pattern of a single neuron is not distinctive and the discriminativeness is distributed among all the neurons. As
empirically observed by \cite{grandmacell}, a classical CNN learns a combination of ``grandmother cells'' and a
distributed code. This observation is further supported by \cite{bolei}, which found that by taking proper weighted
average over all the feature maps produced by a convolutional layer, one can effectively visualize all the regions
in the input image used for classification. Note that both \cite{grandmacell} and \cite{bolei} are based on the original CNN
structure and the quality of representation learning remains the same or slightly worse for the sake of better
localization. On the other hand, \cite{dsn, lcnn, cite_lcnn} learn more discriminative representations by putting 
supervision on intermediate layers, usually by transforming the fully-connected layer output through another fully-connected layer followed by
a loss layer. These transformations introduce a separation between the supervisory signal and internal filters that makes
their methods difficult to
visualize. A more recent related work is the popular SSD \cite{ssd} detection framework; it associates a
convolutional filter with either a particular category of certain aspect ratio or certain location coordinates. Compared
to SSD, our architecture operates at a finer-level (small patches instead of objects) and is optimized for recognition.

\section{Learning Discriminative Patch Detectors as a Bank of Convolutional Filters} \label{sec3}
We regard a $1\times1$ convolutional
filter as a small patch detector. Specifically, referring to Figure \ref{fig1}, if we pass an input
image through a series of convolutional and pooling layers to obtain a feature map of size $C\times H\times W$, 
each $C \times 1 \times 1$ vector across channels at fixed spatial location represents a small
patch at a corresponding location in the original image. Suppose we have learned a $1\times1$ filter which has high
response to a certain discriminative region; by convolving the feature map with this filter we obtain a heatmap.
Therefore, a
discriminative patch can be found simply by picking the location with the maximum value in the entire heatmap. This operation of
spatially pooling the entire feature map into a single value is defined as Global Max Pooling (GMP) \cite{bolei}.

Two requirements are needed to make the feature map suitable for this idea.
First, since the discriminative regions in fine-grained categories are usually highly localized, we need a relatively
small receptive field, \ie, each $C\times1\times1$ vector represents a relatively small patch in the original image.
Second, since fine-grained recognition involves accurate patch localization, the stride in
the original image between adjacent
patches should also be small. In early network architectures, the size and stride of the
convolutional filters and pooling kernels were large. As a result, the receptive
field of a single neuron in later convolutional
layers was large, as was the stride between adjacent fields. Fortunately, the evolution of
network architectures \cite{vgg, googlenet, resnet} has led to smaller filter sizes and pooling kernels. For
example, in a 16-layer VGG network (VGG-16), the output of the $10^{\rm th}$ convolutional layer \texttt{conv4\_3}
represents patches as small as $92\times92$ with stride 8, which is small and dense enough for our task given common
CNN input size.

\begin{figure}
\begin{center}
\includegraphics[width=0.45\textwidth]{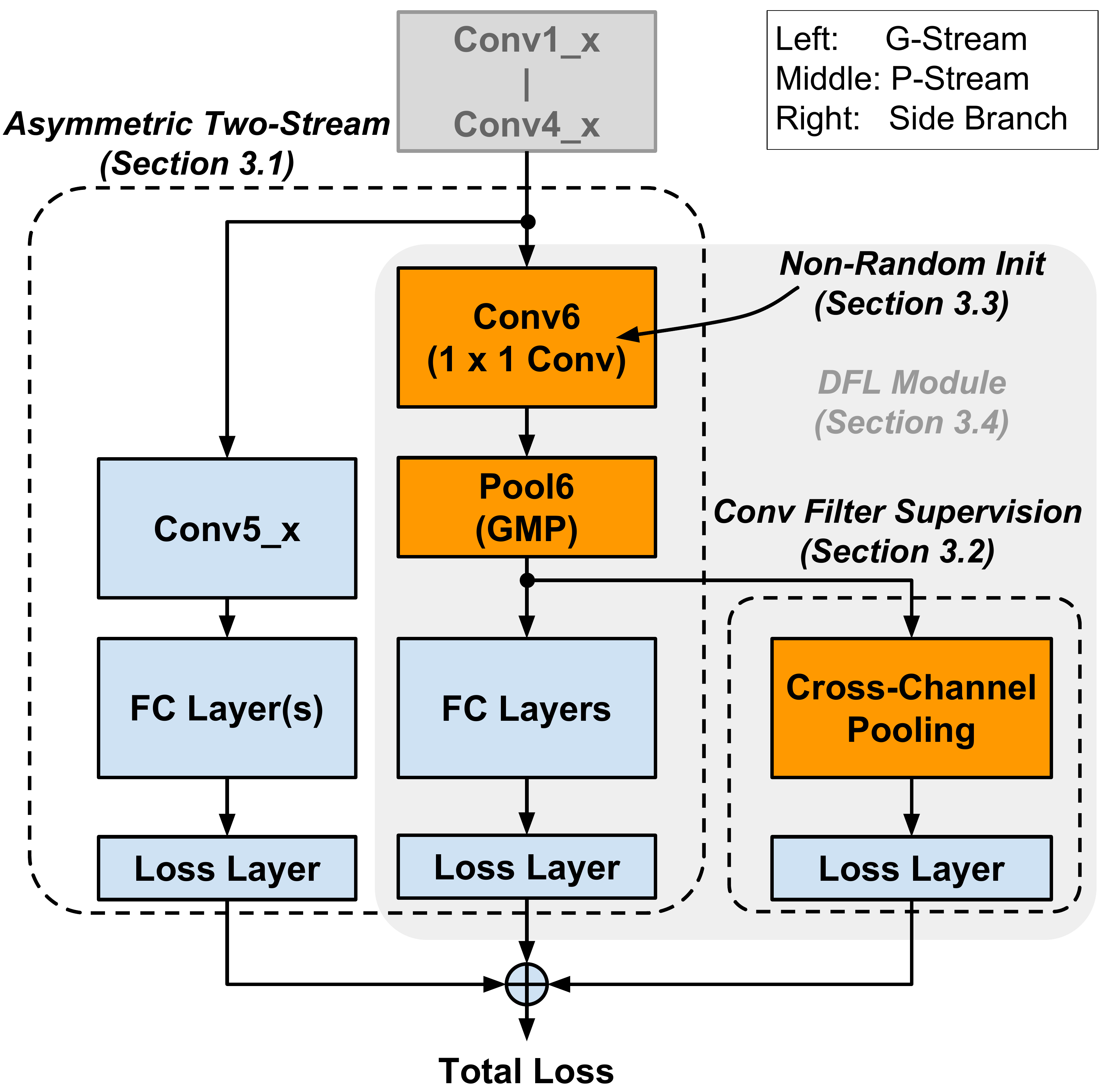}
\end{center}
\vspace{-10pt}
\caption{\label{fig2} Overview of our framework, which consists of a) an asymmetric two-stream architecture to learn
both the discriminative patches and global features, b) supervision imposed to learn discriminative patch detectors and
c) non-random layer initialization. For simplicity, except GMP, all pooling and ReLU layers between
convolutional layers are not displayed.}
\end{figure}

In the rest of Section \ref{sec3}, we demonstrate how a set of discriminative patch detectors can be
learned as a $1\times1$ convolutional layer in a network specifically designed for this task. An overview of our framework is
displayed in Figure \ref{fig2}. There are three key components in our design: an asymmetric
two-stream structure to learn discriminative patches as well as global features (Section \ref{sec3_2}), 
convolutional filter supervision to ensure the discriminativeness of the patch detectors (Section \ref{sec3_3}) and
non-random layer initialization to accelerate the network convergence (Section \ref{sec3_4}). We then extend our
framework to handle patches of different scales (Section \ref{sec3_5}). We use VGG-16 for illustration, but our ideas
are not limited to any specific network architecture as our experiments show.

\subsection{Asymmetric Two-stream Architecture} \label{sec3_2}
The core component of the network responsible for discriminative patch learning is a $1\times1$ convolutional
layer followed by a GMP layer, as displayed in Figure \ref{fig1}. 
This component followed by a classifier (\eg, fully-connected layers and a softmax layer) forms the discriminative patch
stream (P-Stream) of our network, where the prediction is made by inspecting the responses of the discriminative patch
detectors. The P-Stream uses the output of \texttt{conv4\_3} and the minimum receptive field in
this feature map corresponds to a patch of size $92\times92$ with stride 8.

The recognition of some fine-grained categories might also depend on global shape and appearance, so another stream
preserves the further convolutional layers
and fully connected layers, where the neurons in the first fully connected layer encode global information by
linearly combining the whole convolutional feature maps. Since this stream focuses on global features, we refer to it as the G-Stream.

We merge the two streams in the end.

\begin{figure}
\begin{center}
\includegraphics[width=0.49\textwidth]{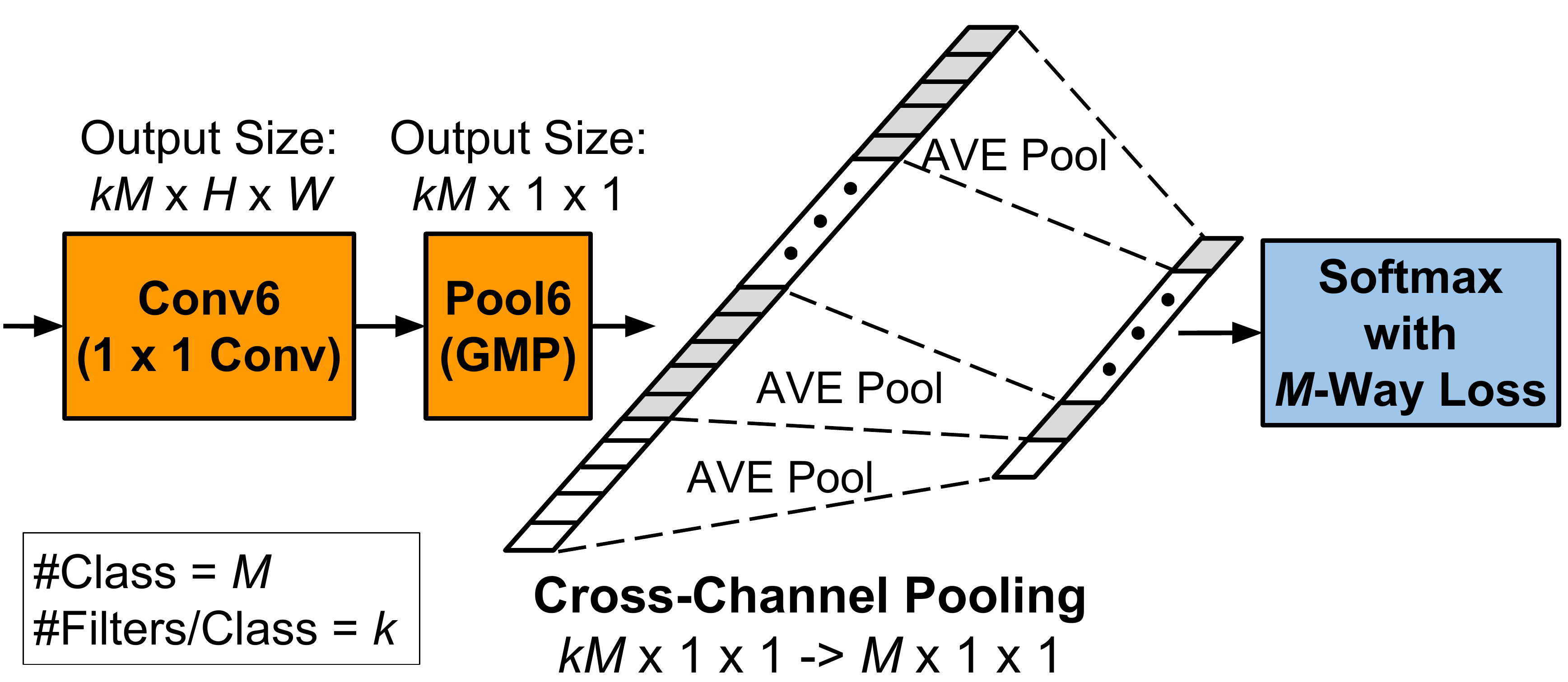}
\end{center}
\vspace{-10pt}
   \caption{\label{fig3} The illustration of our convolutional filter supervision. The filters in \texttt{conv6} are
   grouped into $M$ groups, where $M$ is the number of classes. The maximum responses in group $i$ are averaged
   into a single score indicating the effect of the discriminative patches in Class $i$. The pooled vector is
   fed into a softmax loss layer to encourage discriminative patch learning.}
\end{figure}

\subsection{Convolutional Filter Supervision} \label{sec3_3}
Using the network architecture described above, the $1\times1$ convolutional layer in the P-Stream is not guaranteed to fire at
discriminative patches as desired. For the framework to learn class-specific discriminative patch detectors, we impose supervision
directly at the $1\times1$ filters by introducing a Cross-Channel Pooling layer followed by a softmax loss layer,
shown in Figure \ref{fig3} as part of the whole framework (the side branch) in Figure \ref{fig2}.

Filter supervision works as follows. Suppose we have $M$ classes and each class has $k$ discriminative patch
detectors; then the number of $1\times1$ filters required is $kM$. After obtaining the max response of each filter through
GMP, we get a $kM$-dimensional vector. Cross-Channel Pooling averages the values across every group of $k$
dimensions as the response of a certain class, resulting in an $M$-dimensional vector. By feeding the pooled vector into
an $M$-way softmax loss, we encourage
the filters from any class to find discriminative patches from training samples of that class, such that their
averaged filter response is large. We use average instead of max
pooling to encourage all the filters from a given class to have balanced responses. Average pooling tends
to affect all pooled filters during back propogation, while max pooling only affects the filter with the maximum response.
Similar considerations are discussed in \cite{bolei}.

Since there is no learnable parameter between the softmax loss and the $1\times1$
convolutional layer, we directly adjust the
filter weights via the loss function. 
In contrast, previous approaches which introduce intermediate supervision \cite{dsn, lcnn, cite_lcnn} have learnable
weights (usually a fully-connected layer)
between the side loss and the main network, which learn the weights of a classifier unused at test time. The
main network is only affected by back-propogating the gradients of these weights. We believe this is a key difference
of our approach from previous ones.

\subsection{Layer Initialization} \label{sec3_4}
In practice, if
the $1\times1$ convolutional layer is initialized randomly, with filter supervision it may converge to bad local minima. For
example, the output vector of the Cross-Channel Pooling can 
approach all-zero or some constant to reduce the side loss during training, a degenerate solution. To overcome the
issue, we introduce a method for non-random initialization.

The non-random initialization is motivated by our interpretation of a $1\times1$ filter as a patch detector.
The patch detector of Class $i$ is initialized by patch representations from the samples in
that class, using weak supervion without part annotations. Concretely, a patch is represented by a $C\times 1\times 1$ vector
at corresponding spatial location of the feature map. We extract the
\texttt{conv4\_3} features from the ImageNet pretrained model and compute the
energy at each spatial location ($l_2$ norm of each $C$-dimensional vector in a feature map). As shown in
the first row of Figure \ref{fig7}, though not perfect, the heatmap of energy distribution acts as a reasonable indicator of useful
patches. Then the vectors with high $l_2$ norms are selected via non-maximum suppression with small overlap threshold; $k$-means
is performed over the selected $C$-dimensional vectors within Class $i$ and the
cluster centers are used as the initializations for filters from
Class $i$. To increase their discriminativeness, we further whiten the initializations using \cite{ldadet} and do $l_2$
normalization. In practice this simple method provides reasonable initializations which are further refined during
end-to-end training. Also, in Section \ref{sec4} we show that the energy distribution becomes much more discriminative after training.

As long as the layer is properly initialized, the whole network can be trained in an end-to-end fashion just once, which is
more efficient compared with the multistage training strategy of previous works \cite{deeplac, spda_cnn, partstack}.

\subsection{Extension: Multiple Scales} \label{sec3_5}
Putting Section \ref{sec3_2} to \ref{sec3_4} together, the resulting framework can utilize discriminative patches
from a single scale. A natural and necessary extension is to utilize patches from multiple scales, since in visual
domains such as birds and aircrafts, objects might have larger scale variations.

As discussed in Section \ref{sec3_2}, discriminative patch size depends on the receptive field of
the input feature map. Therefore, multi-scale extension of our approach is equivalent to utilizing multiple feature
maps. We regard the P-Stream and side branch (with non-random initialization) together as a ``Discriminative Filter 
Learning'' (DFL) module that is added after \texttt{conv4\_3} in Figure \ref{fig2}. By simply
adding the DFL modules after multiple convolutional layers we achieve multi-scale patch learning. In practice,
feature maps produced by very early convolutional layers are not suitable for class-specific operations since
they carry information that is too low-level, therefore
the DFL modules are added after several late convolutional layers in Section \ref{sec4}.

Our multi-layer branch-out is inspired by recent approaches in object detection \cite{ssd,
fanyang16}, where feature maps from multiple convolutional layers are directly used to detect objects of 
multiple scales. Compared with these works, our approach operates at a finer level
and is optimized for recognition instead of localization.

\section{Experiments} \label{sec4}
In the rest of this paper, we denote our approach by DFL-CNN, which is an abbreviation for \textit{Discriminative Filter Learning within a
CNN}. We use the following datasets: 

\noindent\textbf{CUB-200-2011} \cite{cub2011} has 11,788 images from 200 classes officially split into 5,994
training and 5,794 test images. 

\noindent\textbf{Stanford Cars} \cite{car196} has 16,185 images from 196 classes officially split into 8,144
training and 8,041 test images.

\noindent\textbf{FGVC-Aircraft} \cite{fgvc_air} has 10,000 images from 100 classes officially split into
 6,667 training and 3,333 test images.
\subsection{Implementation Details} \label{sec4_2}
We first describe the basic settings of our DFL-CNN and then we introduce two higher-capacity
settings. The input size of all our networks is $448\times448$, which is standard in the literature. We do not use part
or bounding box (BBox) annotations and compare our method with other weakly-supervised approaches (without part
annotation). In addition, no model ensemble is used in our experiments.

The network structure of our basic DFL-CNN is based on 16-layer VGGNet \cite{vgg} and the DFL module is added after
\texttt{conv4\_3}, as illustrated exactly in Figure \ref{fig2}.
In \texttt{conv6}, we set the number of filters per class to be 10. During Cross-Channel average
pooling, the maximum responses of each group of 10 filters are pooled into one dimension. At initialization time,
\texttt{conv6} is initialized in the way discussed in Section \ref{sec3_4}; other original VGG-16 layers are initialized
from an ImageNet pretrained model directly (compared with ``indirect'' initialization of \texttt{conv6}) and other newly
introduced layers are randomly initialized. After initialization, a single stage end-to-end training proceeds, with the
G-Stream, P-Stream and side branch having their own softmax with cross-entropy losses with 
weights 1.0, 1.0 and 0.1 respectively. At test time, these softmax-with-loss layers are removed and the prediction is
the weighted combination of the outputs of the three streams.

We extend DFL-CNN in two ways. The first extension, 2-scale DFL-CNN, was discussed in Section \ref{sec3_5}. In
practice, two DFL modules are added after \texttt{conv4\_3} and \texttt{conv5\_2}, while the output of the last
convolutional layer (\texttt{conv5\_3}) is used by G-Stream to extract global information. The second extension
shows that our approach applies to other network architectures, a 50-layer
ResNet \cite{resnet} in this case. Similar to VGGNet, ResNet also groups convolutional layers into five groups and our DFL
module is added to the output of the fourth group (\ie \texttt{conv4\_x} in \cite{resnet}). Initialization,
training and testing of the two extended networks are the same as basic DFL-CNN.

\subsection{Results} \label{sec4_3}
The results on CUB-200-2011, Stanford Cars and FGVC-Aircraft are displayed in Table \ref{tb2}, Table \ref{tb3} and Table
\ref{tb4}, respectively. In each table from top to bottom, the methods are separated into five groups, as discussed in Section \ref{sec1},
which are (1) fine-tuned
baselines, (2) CNN features + multi-stage frameworks, (3) localization-classification subnets,
(4) end-to-end feature encoding and (5) DFL-CNN. The basic DFL-CNN, 2-scale extension and ResNet extension in
Section \ref{sec4_2} are denoted by ``DFL-CNN (1-scale) / VGG-16'', ``DFL-CNN (2-scale) / VGG-16'' and ``DFL-CNN
(1-scale) / ResNet-50'', respectively. Our VGG-16 based approach not only outperforms corresponding fine-tuned baseline by
a large margin, but also achieves or outperforms state-of-the-art under the same base model; our best results further
outperform state-of-the-art by a noticeable margin on all datasets, suggesting its effectiveness.

Earlier multi-stage frameworks built upon CNN features achieve comparable results, while they often require bounding box
annotations and the multi-stage nature limits
their potential. The end-to-end feature encoding methods have very high performance on
birds, while their advantages diminish when dealing with rigid objects. The localization-classification subnets achieve
high performance on various datasets, usually with a large number of network parameters. For instance, the STN
\cite{stn} consists of an Inception localization network followed by four Inception classification networks without
weight-sharing, and RA-CNN \cite{taomei1} consists of three independent VGGNets and two localization sub-networks. 
Our end-to-end approach achieves state-of-the-art with no extra
annotation, enjoys consistent performance on both rigid and non-rigid objects, and has relatively compact network architecture.
\begin{table}
\centering
\resizebox{0.4\textwidth}{!}{
\begin{tabular}{|c|c|c|}
\hline
Method & Base Model & Accuracy (\%)\\
\hline
\hline
FT VGGNet \cite{taomei1} & VGG-19 & 77.8\\
FT ResNet & ResNet-50 & 84.1\\
\hline
CoSeg(+BBox) \cite{krause15} & VGG-19 & 82.6\\
PDFS \cite{deepresp} & VGGNet & 84.5\\
\hline
STN \cite{stn} & Inception \cite{batchnorm} & 84.1\\
RA-CNN \cite{taomei1} & VGG-19 & 85.3 \\
MA-CNN \cite{taomei2} & VGG-19 & \textbf{86.5} \\
\hline
B-CNN \cite{b_cnn} & VGG-16 & 84.1\\
Compact B-CNN \cite{c_b_cnn} & VGG-16 & 84.0\\
Low-rank B-CNN \cite{lowrank_bcnn} & VGG-16 & 84.2\\
Kernel-Activation \cite{highorder2} & VGG-16 & 85.3\\
Kernel-Pooling \cite{highorder1} & VGG-16 & 86.2\\
Kernel-Pooling \cite{highorder1} & ResNet-50 & 84.7\\
\hline

\hline

\hline
DFL-CNN (1-scale)& VGG-16 & 85.8\\
DFL-CNN (2-scale)& VGG-16 & 86.7\\
DFL-CNN (1-scale)& ResNet-50 & \textcolor{red}{\textbf{87.4}}\\
\hline
\end{tabular}}
\vspace{1pt}
\caption{\label{tb2}Comparison of our approach (DFL-CNN) to recent results on
CUB-200-2011, \textbf{without} extra annotations (if not specified). For the
finetuned (FT) baselines, we cite the best previously reported result if it is better than our implementation. The
black-bold number represents the best previous result.}
\vspace{-10pt}
\end{table}

\begin{table}
\centering
\resizebox{0.4\textwidth}{!}{
\begin{tabular}{|c|c|c|}
\hline
Method & Base Model & Accuracy (\%)\\
\hline
\hline
FT VGGNet \cite{taomei1} & VGG-19 & 84.9\\
FT ResNet & ResNet-50 & 91.7\\
\hline
BoT(+BBox) \cite{tripmine} & VGG-16 & 92.5\\
CoSeg(+BBox) \cite{krause15} & VGG-19 & 92.8\\
\hline
RA-CNN \cite{taomei1} & VGG-19 & 92.5\\
MA-CNN \cite{taomei2} & VGG-19 & \textbf{92.8}\\
\hline
B-CNN \cite{b_cnn} & VGG-16 & 91.3\\
Low-Rank B-CNN \cite{lowrank_bcnn} & VGG-16 & 90.9\\
Kernel-Activation \cite{highorder2} & VGG-16 & 91.7\\
Kernel-Pooling \cite{highorder1} & VGG-16 & 92.4\\
Kernel-Pooling \cite{highorder1} & ResNet-50 & 91.1\\
\hline

\hline

\hline
DFL-CNN (1-scale) & VGG-16 & 93.3\\
DFL-CNN (2-scale) & VGG-16 & \textcolor{red}{\textbf{93.8}}\\
DFL-CNN (1-scale) & ResNet-50 & 93.1\\
\hline
\end{tabular}}
\vspace{1pt}
\caption{\label{tb3}Comparison of our approach (DFL-CNN) to recent results on
Stanford Cars \textbf{without} extra annotations (if not specified).}
\vspace{-10pt}
\end{table}

\begin{table}
\centering
\resizebox{0.4\textwidth}{!}{
\begin{tabular}{|c|c|c|}
\hline
Method & Base Model & Accuracy (\%)\\
\hline
\hline
FT VGGNet & VGG-19 & 84.8\\
FT ResNet & ResNet-50 & 88.5\\
\hline
MGD(+BBox) \cite{mgd} & VGG-19 & 86.6\\
BoT(+BBox) \cite{tripmine} & VGG-16 & 88.4\\
\hline
RA-CNN \cite{taomei1} & VGG-19 & 88.2\\
MA-CNN \cite{taomei2} & VGG-19 & \textbf{89.9}\\
\hline
B-CNN \cite{b_cnn} & VGG-16 & 84.1\\
Low-Rank B-CNN \cite{lowrank_bcnn} & VGG-16 & 87.3\\
Kernel-Activation \cite{highorder2} & VGG-16 & 88.3\\
Kernel-Pooling \cite{highorder1} & VGG-16 & 86.9\\
Kernel-Pooling \cite{highorder1} & ResNet-50 & 85.7\\
\hline

\hline
DFL-CNN (1-scale) & VGG-16 & 91.1\\
DFL-CNN (2-scale) & VGG-16 & \textcolor{red}{\textbf{92.0}}\\
DFL-CNN (1-scale) & ResNet-50 & 91.7\\
\hline
\end{tabular}}
\vspace{1pt}
\caption{\label{tb4}Comparison of our approach (DFL-CNN) to recent results on
FGVC-Aircraft \textbf{without} extra annotation (if not specified).}
\vspace{-10pt}
\end{table}

Our approach can be applied to various network architectures.
Most previous approaches in fine-grained recognition have based their network on VGGNets and previously reported
ResNet-based results are less effective than VGG-based ones. Table \ref{tb2}, \ref{tb3} and \ref{tb4} shows that our
ResNet baseline is already very
strong, however our ResNet based DFL-CNN is able to outperform the strong baseline by a large margin (\eg 3.3\% absolute
percentage on birds). This clearly indicates that CNN's mid-level learning capability can still be improved even
though the network is very deep.

\subsection{Ablation Studies} \label{sec4_5}
We conduct ablation studies to understand the components of our approach. These
experiments use the basic DFL-CNN framework and the CUB-200-2011 dataset.

\noindent\textbf{Contribution of each stream}\quad Given a trained DFL-CNN, we investigate the contribution of each
stream at test time. Table \ref{tb5} shows that the performance of the G-Stream or
P-Stream alone is mediocre, but the combination of the two is significantly better than either one alone,
indicating that the global information and the discriminative patch information are highly complementary. Additionally, 
the side branch provides extra gain to reach the full performance in Table \ref{tb2}.
\begin{table}
\centering
\resizebox{0.25\textwidth}{!}{
\begin{tabular}{|c|c|}
\hline
Settings & Accuracy (\%)\\
\hline
\hline
G-Stream Only & 80.3\\
P-Stream Only & 82.0\\
G + P & 84.9\\
G + P + Side & 85.8\\
\hline
\end{tabular}}
\vspace{1pt}
\caption{\label{tb5}Contribution of the streams \textit{at test time} on CUB-200-2011. Note that at training time a \textit{full} DFL-CNN model is
trained, but the prediction only uses certain stream(s).}
\vspace{-10pt}
\end{table}

\begin{table}
\centering
\resizebox{0.25\textwidth}{!}{
\begin{tabular}{|c|c|}
\hline
\texttt{pool6} Method & Accuracy (\%)\\
\hline
\hline
GMP & 85.8\\
GAP & 80.4\\
\hline
\end{tabular}}
\vspace{1pt}
\caption{\label{tb7}Effect of Global Max Pooling (GMP) vs. Global Average Pooling (GAP) on CUB-200-2011.}
\vspace{-10pt}
\end{table}

\begin{table}
\centering
\resizebox{0.4\textwidth}{!}{
\begin{tabular}{|c|c|c|}
\hline
Layer Initialization & Filter Supervision & Accuracy (\%)\\
\hline
\hline
- & - & 82.2\\
\checkmark & - & 84.4\\
\checkmark & \checkmark & 85.8\\
\hline
\end{tabular}}
\vspace{1pt}
\caption{\label{tb6}Effect of intermediate supervision of DFL-CNN \textit{at training time}, evaluated on CUB-200-2011.}
\vspace{-10pt}
\end{table}

\begin{table}
\centering
\resizebox{0.4\textwidth}{!}{
\begin{tabular}{|c|c|c|}
\hline
Method & Without BBox (\%) & With BBox (\%)\\
\hline
\hline
FT VGG-16 \cite{fzhou1} & 74.5 & 79.8 \\
DFL-CNN & 85.8 & 85.7\\
\hline
\end{tabular}}
\vspace{1pt}
\caption{\label{tb8}Effect of BBox evaluated on CUB-200-2011.}
\vspace{-10pt}
\end{table}

\noindent\textbf{Effect of intermediate supervision}\quad We investigate the effect of Section
\ref{sec3_3} and \ref{sec3_4} by training the DFL-CNN without certain component(s) and comparing with the full model.
Table \ref{tb6} shows a significant performance improvement when we gradually add the intermediate supervision
components to improve the quality of learned discriminative filters. Note that Table \ref{tb6} does not include ``Filter
Supervision without Layer Initialization'' settings since it leads to failure to converge of P-Stream as mentioned in Section
\ref{sec3_4}.

\begin{figure*}
\begin{center}
\includegraphics[width=\textwidth]{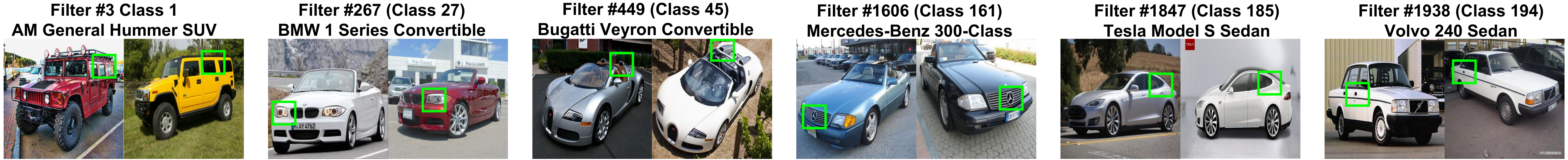}
\end{center}
   \vspace{-10pt}
\caption{\label{fig5} The visualization of top patches in Stanford Cars. We remap the spatial location of the highest
activation in a feature map back to the patch in the original image. The results are highly consistent
with human perception, and cover diverse regions such as head light
(\textbf{$2^{\rm nd}$ column}), air intake (\textbf{$3^{\rm th}$ column}), frontal face (\textbf{$4^{\rm th}$ column})
and the black side stripe (\textbf{last column}).}
   \vspace{-10pt}
\end{figure*}

\begin{figure*}
\begin{minipage}{\textwidth}
\includegraphics[width=\textwidth]{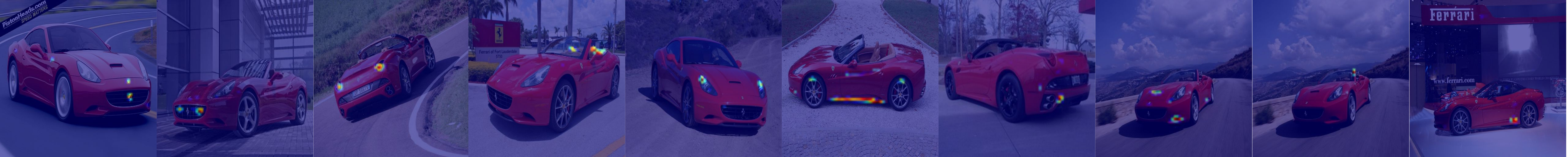}
\caption{\label{fig11} Sample visualization of all ten filter activations learned for one class (Class
102) by upsampling the \texttt{conv6} feature maps to image resolution, similar to \cite{bolei}. The activations are 
disriminatively concentrated and cover diverse regions. Better viewed at 600\%.}
\end{minipage}
\begin{minipage}{\textwidth}
\includegraphics[width=0.33\textwidth]{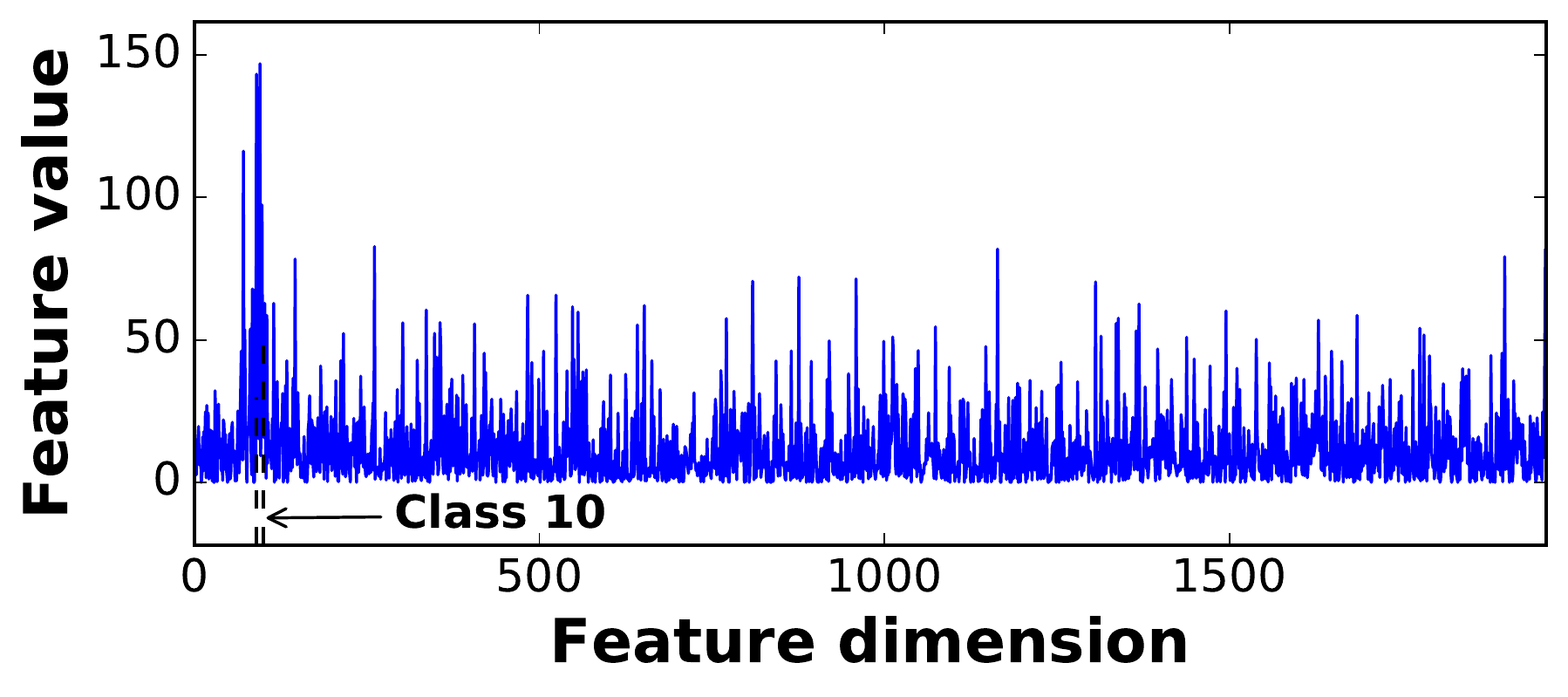}
\includegraphics[width=0.33\textwidth]{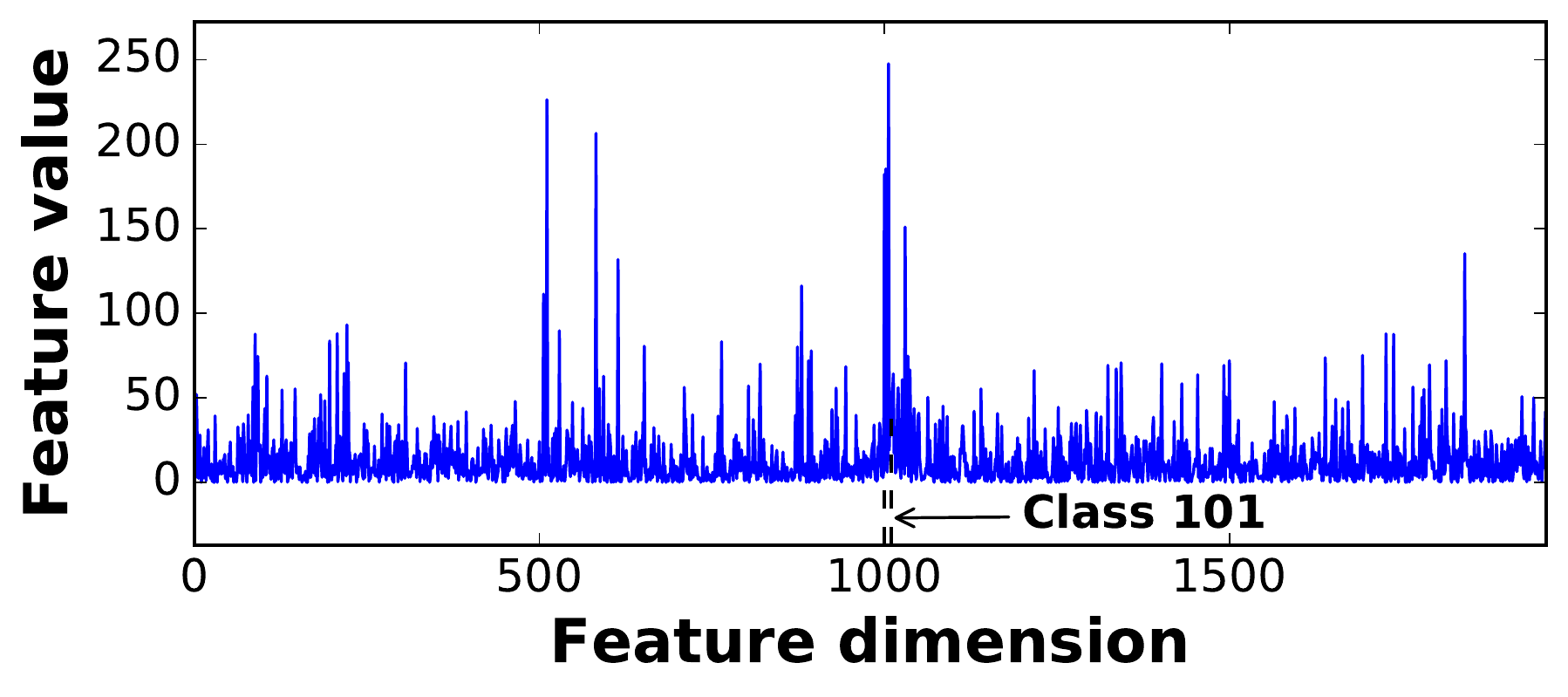}
\includegraphics[width=0.33\textwidth]{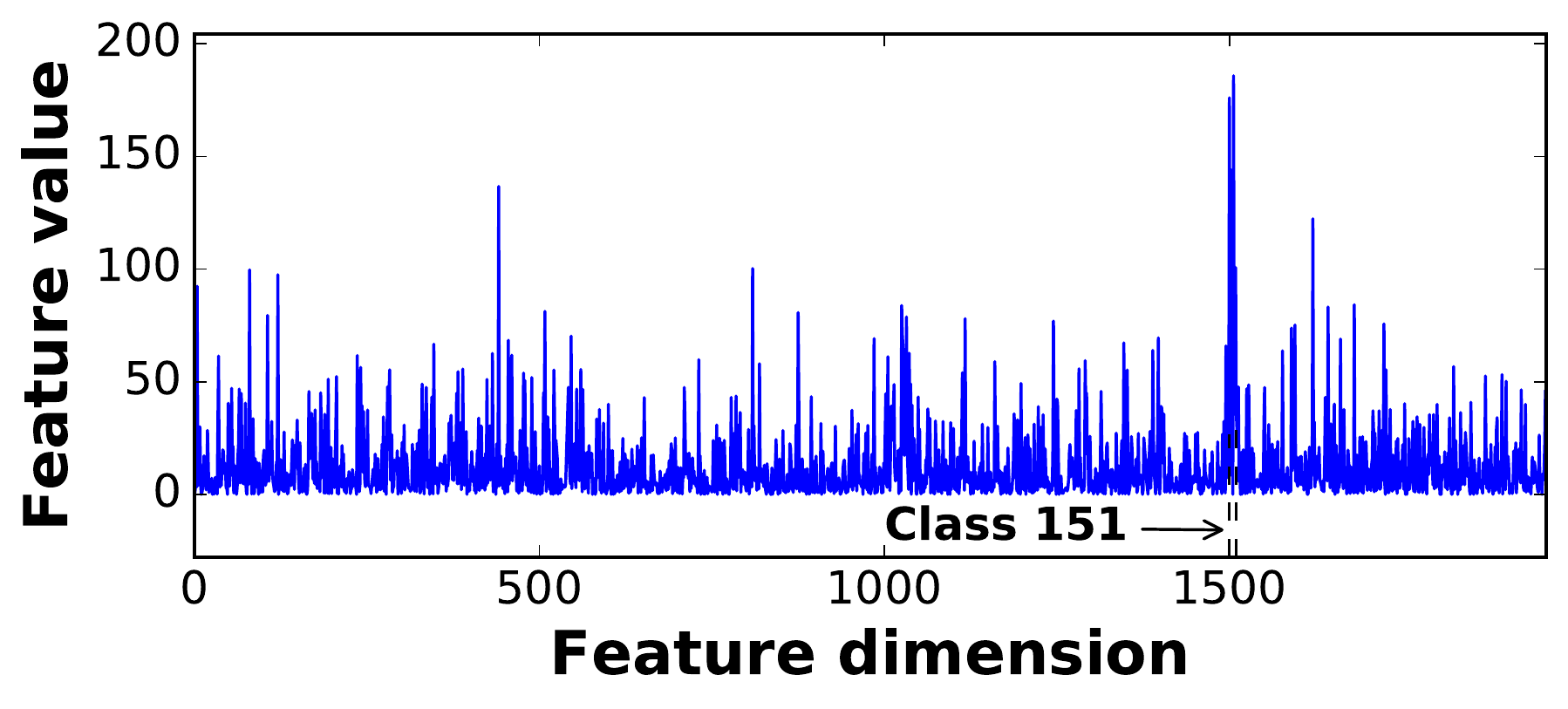}
   \vspace{-10pt}
\caption{\label{fig6} The \texttt{pool6} features averaged over all test samples from Class 10,
101 and 151 in Stanford Cars. The dash lines indicate the range of values given by the discriminative patch detectors belonging to the
class. The representations peak at the corresponding class.}
   \vspace{-10pt}
\end{minipage}
\end{figure*}
\noindent\textbf{GMP vs. GAP}\quad More insight into the training process can be obtained by simply switching the pooling method
of \texttt{pool6} in Figure \ref{fig2}. As can be seen from the Table \ref{tb7}, switching the pooling method from
GMP to Global Average Pooling (GAP) leads to a significant performance drop such that the accuracy is
close to ``G-Stream Only'' in Table \ref{tb5}. Therefore, although
\texttt{conv6} is initialized to the same state, during training GMP makes the filters more discriminative by
encouraging the $1\times1$ filters to have very high response at a certain location of the feature map and the gradients
will only be back-propagated to that location, while GAP makes the P-Stream almost useless by encouraging the filters
to have mediocre responses over the whole feature maps and the gradients affect every spatial location.

\noindent\textbf{Unnecessary BBox.}\quad Since our approach, DFL-CNN, is able to utilize discriminative patches without
localization, it is expected to be less sensitive to BBox than the fine-tuned baseline, as supported by the
results in Table \ref{tb8}.

\subsection{Visualization and Analysis} \label{sec4_4}
Insights into the behavior of our approach can be obtained by visualizing the effects of \texttt{conv6}, the $1\times1$ convolutional layer. To 
understand its behavior, we
\begin{itemize}[noitemsep, topsep=0pt]
\item visualize patch activations. Since we regard each filter as a discriminative patch detector, we identify
the learned patches by remapping spatial locations of top filter activations back to images. Figure \ref{fig5} shows
that we do find high-quality discriminative regions.
\item visualize a forward pass. Since the max responses of these filters are directly used for
classification, by visualizing the output of \texttt{conv6}'s next layer, \texttt{pool6}, we find that it
produces discriminative representations which have high responses for certain classes.
\item visualize back propagation. During training, \texttt{conv6} can affect its previous layer, \texttt{conv4\_3}
(VGG-16), through back propagation. By comparing the \texttt{conv4\_3} features before and after training, we find that the
spatial energy distributions of previous feature maps are changed in a discriminative fashion. 
\end{itemize}
\begin{figure}
\begin{center}
\includegraphics[width=0.4\textwidth]{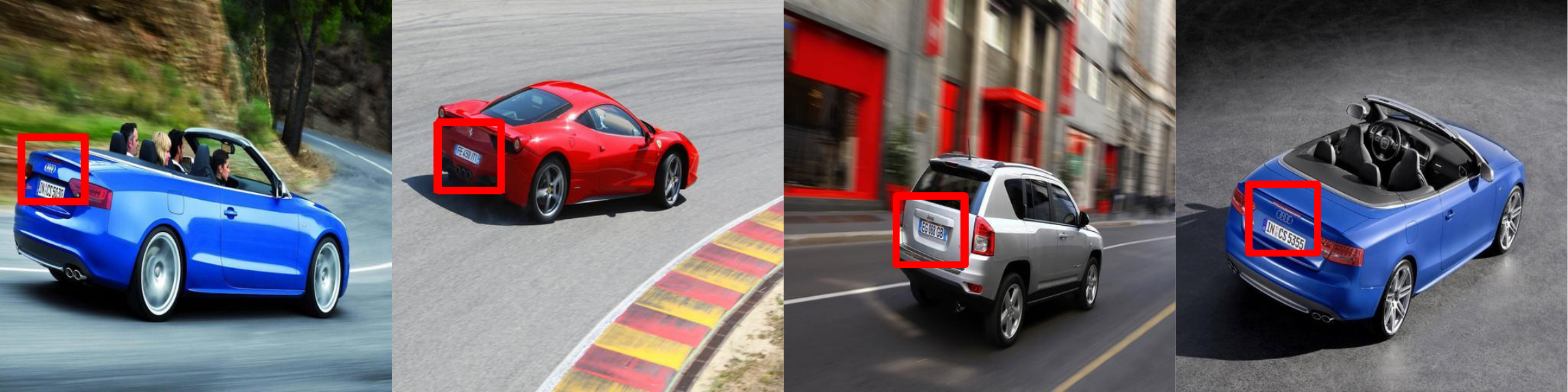}
\end{center}
   \vspace{-10pt}
   \caption{\label{fig12} Visualization of a failure case, where the filter activates on commonly appeared licence plates.}
   \vspace{-10pt}
\end{figure}

\begin{figure}
\begin{center}
\includegraphics[width=0.4\textwidth]{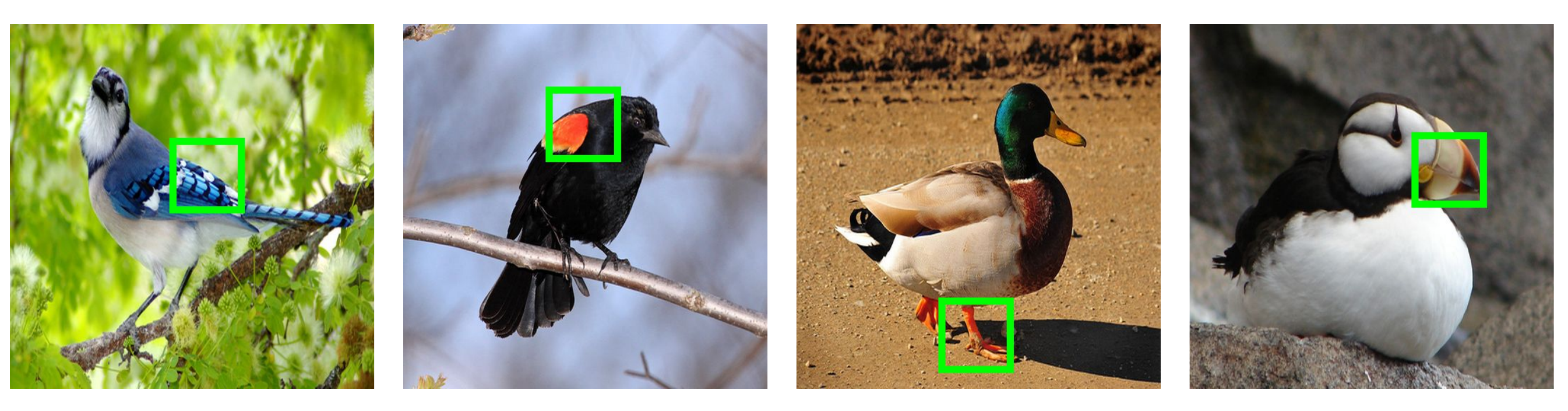}
\end{center}
   \vspace{-10pt}
   \caption{\label{fig8} The visualization of patches in CUB-200-2011. We accurately
   localize discriminative patches without part annotations, such as the bright texture (\textbf{first image}), the color spot 
   (\textbf{second image}), the webbing and beak (\textbf{third} and \textbf{forth} image).}
   \vspace{-10pt}
\end{figure}

\begin{figure}
\begin{center}
\includegraphics[width=0.4\textwidth]{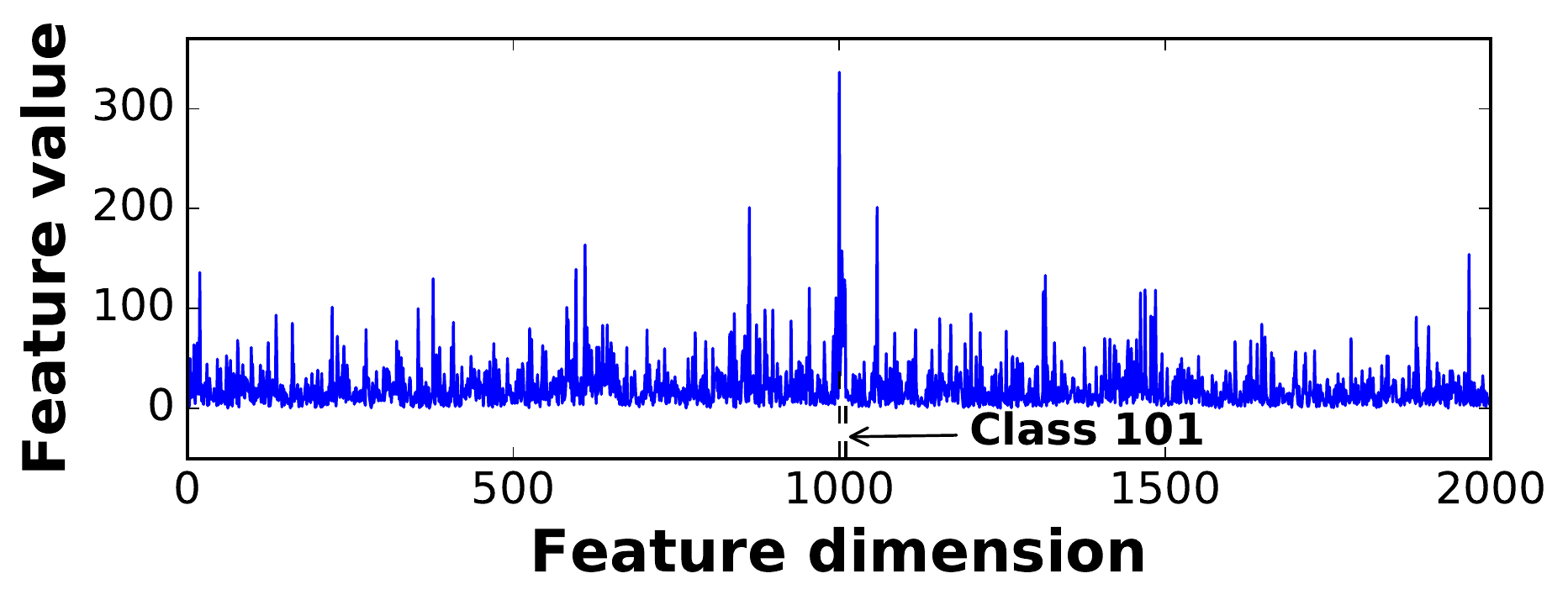}
\end{center}
   \vspace{-10pt}
   \caption{\label{fig9}The averaged \texttt{pool6} features over all test samples from Class 101 in CUB-200-2011, peaky at
   corresponding dimensions.}
   \vspace{-10pt}
\end{figure}

\begin{figure*}
\begin{center}
\includegraphics[width=0.8\textwidth]{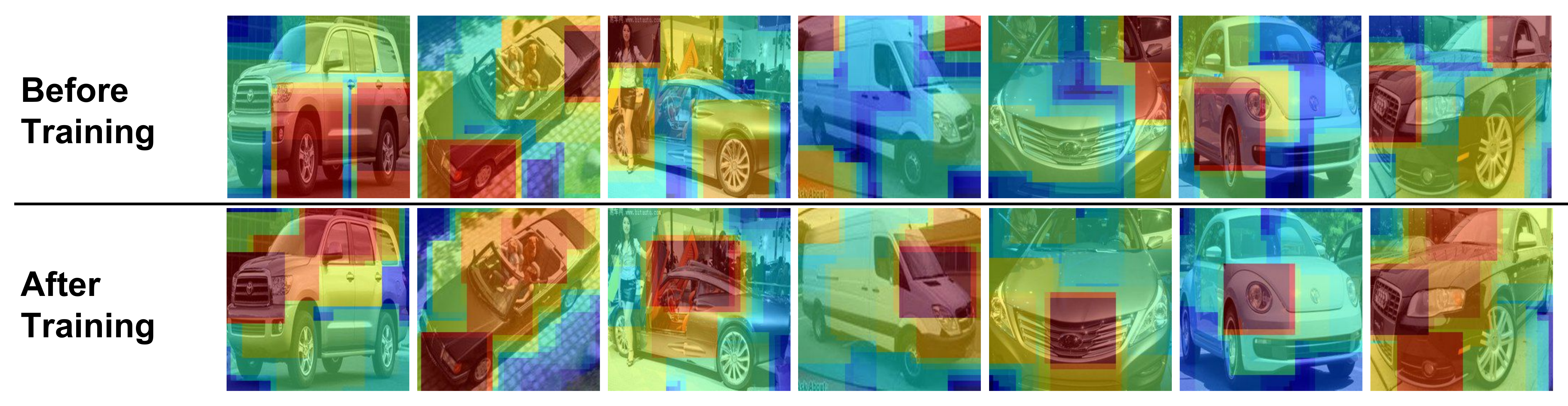}
\end{center}
   \vspace{-10pt}
\caption{\label{fig7}Visualization of the energy distribution of \texttt{conv4\_3} feature map before and after
training for Stanford Cars. We remap each spatial location in the feature map back to the patch in the original image. After training in
our approach, the energy distribution becomes more discriminative. For example, in the \textbf{$1^{\rm st}$ column},
the high energy region shifts from the wheels to discriminative regions like the frontal face and the top of the
vehicle; in the \textbf{$2^{\rm nd}$
column}, after training the energy over the brick patterns is reduced; in the \textbf{$3^{\rm rd}$ column}, the person no longer
lies in high energy region after training; in the \textbf{$7^{\rm th}$
column}, before training the energy is focused mostly at the air grill, and training adds the
discriminative fog light into the high energy region. More examples are
interpretated in Section \ref{sec4_4_2}.}
   \vspace{-10pt}
\end{figure*}

\subsubsection{Stanford Cars} \label{sec4_4_2}
The visualization of top patches found by some classes' $1\times1$ filters is displayed in Figure \ref{fig5}; the
visualization of all ten filters learned for a sample class is displayed in Figure \ref{fig11}. Unlike
previous filter visualizations, which pick human interpretable results randomly among the filter activations, we have
imposed supervision on \texttt{conv6} filters and can identify their corresponding classes. Figure
\ref{fig5} shows that the top patches are very consistent with human perception.
For instance, the $1847^{\rm th}$ filter belonging to
Class 185 (Tesla Model S) captures the distinctive tail of this type. Figure \ref{fig11} shows that the filter
activation are highly concentrated at discriminative regions and the ten filters cover diverse regions.
The network can localize these subtle discriminative regions because:
a) $1\times1$ filters correspond to small patch detectors in original image, b) the filter supervision, and c) the
use of cluster centers as initialization promotes diversity.
\begin{figure}
\begin{center}
\includegraphics[width=0.4\textwidth]{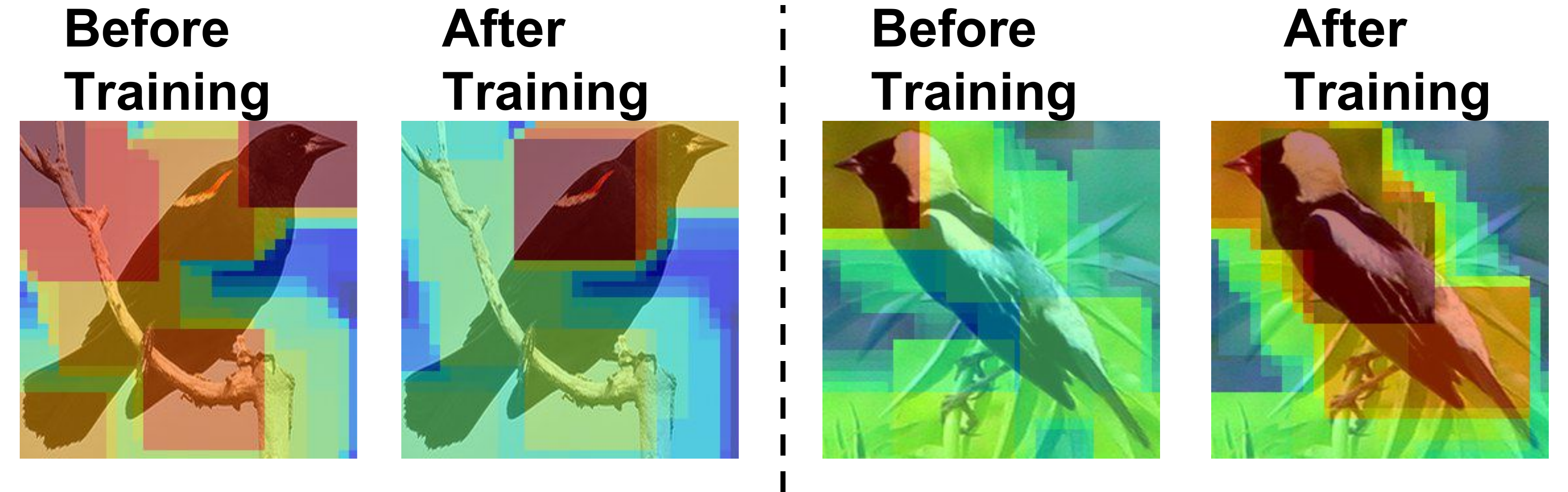}
\end{center}
   \vspace{-10pt}
   \caption{\label{fig10}The energy distributions of \texttt{conv4\_3} feature maps before and after training in
   CUB-200-2011. After training, in the left example, the high energy region at the background branches is greatly
   shrinked and the energy is concentrated at the discriminative color spot; in the right example, more energy is distributed to the
   distinctive black-and-white wing and tail of the species.}
   \vspace{-10pt}
\end{figure}

The visualization of \texttt{pool6} features is shown in Figure \ref{fig6}. We plot the averaged representations over all
test samples from a certain class. Since we have learned a set of discriminative filters, the representations
should have high responses at one class or only a few classes. Figure \ref{fig6} shows that our approach works as
expected. As noticeable, the fine-grained similarity at patch-level (\eg Audi A4 and Audi A6) and few common patterns (
example shown in Figure \ref{fig12}) might explain the alternative peaks in Figure \ref{fig6}.

Most interesting is the effect of \texttt{conv6} on the previous convolutional layer \texttt{conv4\_3} through back
propagation. As discussed in Section \ref{sec3_4},
we use the energy distribution of \texttt{conv4\_3} as a hint to provide layer initialization. After training, we
observed that the energy distribution is refined by \texttt{conv6} and becomes more discriminative, as shown by Figure
\ref{fig7}.
We map every spatial location in the
feature map back to the corresponding patch in the original image, and the value of each pixel is determined by the max
energy patch covering that pixel. From the first line of Figure \ref{fig7}, the features extracted from an ImageNet
pretrained model tend to have high energy at round patterns such as wheels, some unrelated background shape, a person in
the image and some texture patterns, which are common patterns in generic models found in \cite{fergus14}. After
training, the energy shifts from these patterns to discriminative regions of cars. For example, in the $6^{\rm th}$
column, the feature map has high energy initially at both the wheel and the head light; after training, the network
has determined that a discriminative patch for that class (Volkswagen Beetle) is the head light rather than the wheels.
Therefore, \texttt{conv6} have beneficial effects on their previous layer during training.

\subsubsection{CUB-200-2011} \label{sec4_4_3}
Figure \ref{fig8} shows examples of the discriminative patches found by our approach. They include the
texture and spots with bright color as well as specific shape of beak or webbing. Compared with
visualizations of previous works not using part annotations (\eg \cite{krause15, b_cnn}), our approach 
localizes such patches more accurately because our patch detectors operate over denser and smaller
patches and do not have to be shared across categories. 

Similar to cars, features from the next GMP layers are peaky at certain categories (Fig. \ref{fig9}). The energy
distributions of previous convolutional features are also improved: high energy at background regions like 
branches is reduced and the discriminative regions become more focused or diverse according to different categories
(Fig. \ref{fig10}).

\section{Conclusion} \label{sec5}
We have presented an approach to fine-grained recognition based on learning a discriminative filter bank within a CNN
framework in an end-to-end fashion without extra annotation. This is done via an asymmetric multi-stream network
structure with convolutional layer supervision and non-random layer initialization.
Our approach learns high-quality discriminative patches and obtains state-of-the-art performance
on both rigid / non-rigid fine-grained datasets. 

{\small
\bibliographystyle{ieee}
\bibliography{egbib}

\begin{thebibliography}{10}\itemsep=-1pt

\bibitem{grandmacell}
P.~Agrawal, R.~B. Girshick, and J.~Malik.
\newblock Analyzing the performance of multilayer neural networks for object
  recognition.
\newblock In {\em ECCV}, 2014.

\bibitem{embed_fg}
Z.~Akata, S.~E. Reed, D.~Walter, H.~Lee, and B.~Schiele.
\newblock Evaluation of output embeddings for fine-grained image
  classification.
\newblock In {\em CVPR}, 2015.

\bibitem{berg1}
T.~Berg and P.~N. Belhumeur.
\newblock {POOF:} part-based one-vs.-one features for fine-grained
  categorization, face verification, and attribute estimation.
\newblock In {\em CVPR}, 2013.

\bibitem{human2}
S.~Branson, C.~Wah, F.~Schroff, B.~Babenko, P.~Welinder, P.~Perona, and
  S.~Belongie.
\newblock Visual recognition with humans in the loop.
\newblock In {\em ECCV}, 2010.

\bibitem{highorder2}
S.~Cai, W.~Zuo, and L.~Zhang.
\newblock Higher-order integration of hierarchical convolutional activations
  for fine-grained visual categorization.
\newblock In {\em ICCV}, 2017.

\bibitem{symbiotic}
Y.~Chai, V.~S. Lempitsky, and A.~Zisserman.
\newblock Symbiotic segmentation and part localization for fine-grained
  categorization.
\newblock In {\em ICCV}, 2013.

\bibitem{human3}
Y.~Cui, F.~Zhou, Y.~Lin, and S.~Belongie.
\newblock Fine-grained categorization and dataset bootstrapping using deep
  metric learning with humans in the loop.
\newblock In {\em CVPR}, 2016.

\bibitem{highorder1}
Y.~Cui, F.~Zhou, J.~Wang, X.~Liu, Y.~Lin, and S.~Belongie.
\newblock Kernel pooling for convolutional neural networks.
\newblock In {\em CVPR}, 2017.

\bibitem{taomei1}
J.~Fu, H.~Zheng, and T.~Mei.
\newblock Look closer to see better: Recurrent attention convolutional neural
  network for fine-grained image recognition.
\newblock In {\em CVPR}, 2017.

\bibitem{c_b_cnn}
Y.~Gao, O.~Beijbom, N.~Zhang, and T.~Darrell.
\newblock Compact bilinear pooling.
\newblock In {\em CVPR}, 2016.

\bibitem{fast_rcnn}
R.~B. Girshick.
\newblock Fast {R-CNN}.
\newblock In {\em ICCV}, 2015.

\bibitem{rcnn}
R.~B. Girshick, J.~Donahue, T.~Darrell, and J.~Malik.
\newblock Rich feature hierarchies for accurate object detection and semantic
  segmentation.
\newblock In {\em CVPR}, 2014.

\bibitem{fgvcfisher}
P.~H. Gosselin, N.~Murray, H.~J{\'{e}}gou, and F.~Perronnin.
\newblock Revisiting the fisher vector for fine-grained classification.
\newblock {\em Pattern Recognition Letters}, 49:92--98, 2014.

\bibitem{ldadet}
B.~Hariharan, J.~Malik, and D.~Ramanan.
\newblock Discriminative decorrelation for clustering and classification.
\newblock In {\em ECCV}, 2012.

\bibitem{resnet}
K.~He, X.~Zhang, S.~Ren, and J.~Sun.
\newblock Deep residual learning for image recognition.
\newblock In {\em CVPR}, 2016.

\bibitem{lan_fgvc}
X.~He and Y.~Peng.
\newblock Fine-grained image classification via combining vision and language.
\newblock In {\em CVPR}, 2017.

\bibitem{vegfru}
S.~Hou, Y.~Feng, and Z.~Wang.
\newblock Vegfru: A domain-specific dataset for fine-grained visual
  categorization.
\newblock In {\em ICCV}, 2017.

\bibitem{partstack}
S.~Huang, Z.~Xu, D.~Tao, and Y.~Zhang.
\newblock Part-stacked cnn for fine-grained visual categorization.
\newblock In {\em CVPR}, 2016.

\bibitem{batchnorm}
S.~Ioffe and C.~Szegedy.
\newblock Batch normalization: Accelerating deep network training by reducing
  internal covariate shift.
\newblock In {\em ICML}, 2015.

\bibitem{stn}
M.~Jaderberg, K.~Simonyan, A.~Zisserman, and K.~Kavukcuoglu.
\newblock Spatial transformer networks.
\newblock In {\em NIPS}, 2015.

\bibitem{lcnn}
Z.~Jiang, Y.~Wang, L.~S. Davis, W.~Andrews, and V.~Rozgic.
\newblock Learning discriminative features via label consistent neural network.
\newblock In {\em WACV}, 2017.

\bibitem{cite_lcnn}
X.~Jin, Y.~Chen, J.~Dong, J.~Feng, and S.~Yan.
\newblock Collaborative layer-wise discriminative learning in deep neural
  networks.
\newblock In {\em ECCV}, 2016.

\bibitem{lowrank_bcnn}
S.~Kong and C.~Fowlkes.
\newblock Low-rank bilinear pooling for fine-grained classification.
\newblock In {\em CVPR}, 2017.

\bibitem{krause15}
J.~Krause, H.~Jin, J.~Yang, and F.~Li.
\newblock Fine-grained recognition without part annotations.
\newblock In {\em CVPR}, 2015.

\bibitem{unreason}
J.~Krause, B.~Sapp, A.~Howard, H.~Zhou, A.~Toshev, T.~Duerig, J.~Philbin, and
  L.~Fei{-}Fei.
\newblock The unreasonable effectiveness of noisy data for fine-grained
  recognition.
\newblock In {\em ECCV}, 2016.

\bibitem{car196}
J.~Krause, M.~Stark, J.~Deng, and L.~Fei-Fei.
\newblock 3d object representation for fine-grained categorization.
\newblock In {\em International IEEE Workshop on 3D Representation and
  Recognition}, 2013.

\bibitem{hsnet}
M.~Lam, B.~Mahasseni, and S.~Todorovic.
\newblock Fine-grained recognition as hsnet search for informative image parts.
\newblock In {\em CVPR}, 2017.

\bibitem{dsn}
C.~Lee, S.~Xie, P.~W. Gallagher, Z.~Zhang, and Z.~Tu.
\newblock Deeply-supervised nets.
\newblock In {\em AISTATS}, 2015.

\bibitem{deeplac}
D.~Lin, X.~Shen, C.~Lu, and J.~Jia.
\newblock Deep {LAC:} deep localization, alignment and classification for
  fine-grained recognition.
\newblock In {\em CVPR}, 2015.

\bibitem{b_cnn}
T.~Lin, A.~RoyChowdhury, and S.~Maji.
\newblock Bilinear {CNN} models for fine-grained visual recognition.
\newblock In {\em ICCV}, 2015.

\bibitem{yenliang}
Y.~Lin, V.~I. Morariu, W.~H. Hsu, and L.~S. Davis.
\newblock Jointly optimizing 3d model fitting and fine-grained classification.
\newblock In {\em ECCV}, 2014.

\bibitem{ssd}
W.~Liu, D.~Anguelov, D.~Erhan, C.~Szegedy, S.~E. Reed, C.~Fu, and A.~C. Berg.
\newblock {SSD:} single shot multibox detector.
\newblock In {\em ECCV}, 2016.

\bibitem{fcn}
J.~Long, E.~Shelhamer, and T.~Darrell.
\newblock Fully convolutional networks for semantic segmentation.
\newblock In {\em CVPR}, pages 3431--3440, 2015.

\bibitem{fgvc_air}
S.~Maji, E.~Rahtu, J.~Kannala, M.~B. Blaschko, and A.~Vedaldi.
\newblock Fine-grained visual classification of aircraft.
\newblock {\em CoRR}, abs/1306.5151, 2013.

\bibitem{neural_act}
M.~Simon and E.~Rodner.
\newblock Neural activation constellations: Unsupervised part model discovery
  with convolutional networks.
\newblock In {\em ICCV}, 2015.

\bibitem{vgg}
K.~Simonyan and A.~Zisserman.
\newblock Very deep convolutional networks for large-scale image recognition.
\newblock {\em CoRR}, abs/1409.1556, 2014.

\bibitem{boxcar}
J.~Sochor, A.~Herout, and J.~Havel.
\newblock Boxcars: 3d boxes as cnn input for improved fine-grained vehicle
  recognition.
\newblock In {\em CVPR}, 2016.

\bibitem{googlenet}
C.~Szegedy, W.~Liu, Y.~Jia, P.~Sermanet, S.~E. Reed, D.~Anguelov, D.~Erhan,
  V.~Vanhoucke, and A.~Rabinovich.
\newblock Going deeper with convolutions.
\newblock In {\em CVPR}, 2015.

\bibitem{cub2011}
C.~Wah, S.~Branson, P.~Welinder, P.~Perona, and S.~Belongie.
\newblock The caltech-ucsd birds 200-2011 dataset.
\newblock In {\em Technical Report CNS-TR-2011-001, Caltech,}, 2011.

\bibitem{human1}
C.~Wah, G.~V. Horn, S.~Branson, S.~Maji, P.~Perona, and S.~Belongie.
\newblock Similarity comparisons for interactive fine-grained categorization.
\newblock In {\em CVPR}, 2014.

\bibitem{mgd}
D.~Wang, Z.~Shen, J.~Shao, W.~Zhang, X.~Xue, and Z.~Zhang.
\newblock Multiple granularity descriptors for fine-grained categorization.
\newblock In {\em ICCV}, 2015.

\bibitem{tripmine}
Y.~Wang, J.~Choi, V.~Morariu, and L.~S. Davis.
\newblock Mining discriminative triplets of patches for fine-grained
  classification.
\newblock In {\em CVPR}, 2016.

\bibitem{maskcnn}
X.~Wei, C.~Xie, and J.~Wu.
\newblock Mask-cnn: Localizing parts and selecting descriptors for fine-grained
  image recognition.
\newblock {\em CoRR}, abs/1605.06878, 2016.

\bibitem{2attention}
T.~Xiao, Y.~Xu, K.~Yang, J.~Zhang, Y.~Peng, and Z.~Zhang.
\newblock The application of two-level attention models in deep convolutional
  neural network for fine-grained image classification.
\newblock In {\em CVPR}, 2015.

\bibitem{har}
S.~Xie, T.~Yang, X.~Wang, and Y.~Lin.
\newblock Hyper-class augmented and regularized deep learning for fine-grained
  image classification.
\newblock In {\em CVPR}, 2015.

\bibitem{fanyang16}
F.~Yang, W.~Choi, and Y.~Lin.
\newblock Exploit all the layers: Fast and accurate cnn object detector with
  scale dependent pooling and cascaded rejection classifiers.
\newblock In {\em CVPR}, 2016.

\bibitem{large_car}
L.~Yang, P.~Luo, C.~C. Loy, and X.~Tang.
\newblock A large-scale car dataset for fine-grained categorization and
  verification.
\newblock In {\em CVPR}, 2015.

\bibitem{bangpeng}
B.~Yao, G.~R. Bradski, and L.~Fei{-}Fei.
\newblock A codebook-free and annotation-free approach for fine-grained image
  categorization.
\newblock In {\em CVPR}, 2012.

\bibitem{fergus14}
M.~D. Zeiler and R.~Fergus.
\newblock Visualizing and understanding convolutional networks.
\newblock In {\em ECCV}, 2014.

\bibitem{spda_cnn}
H.~Zhang, T.~Xu, M.~Elhoseiny, X.~Huang, S.~Zhang, A.~Elgammal, and D.~Metaxas.
\newblock Spda-cnn: Unifying semantic part detection and abstraction for
  fine-grained recognition.
\newblock In {\em CVPR}, 2016.

\bibitem{fg_rcnn}
N.~Zhang, J.~Donahue, R.~B. Girshick, and T.~Darrell.
\newblock Part-based r-cnns for fine-grained category detection.
\newblock In {\em ECCV}, 2014.

\bibitem{ningzhang3}
N.~Zhang, R.~Farrell, F.~N. Iandola, and T.~Darrell.
\newblock Deformable part descriptors for fine-grained recognition and
  attribute prediction.
\newblock In {\em ICCV}, 2013.

\bibitem{deepresp}
X.~Zhang, H.~Xiong, W.~Zhou, W.~Lin, and Q.~Tian.
\newblock Picking deep filter responses for fine-grained image recognition.
\newblock In {\em CVPR}, 2016.

\bibitem{fzhou2}
X.~Zhang, F.~Zhou, Y.~Lin, and S.~Zhang.
\newblock Embedding label structures for fine-grained feature representation.
\newblock In {\em CVPR}, 2016.

\bibitem{taomei2}
H.~Zheng, J.~Fu, T.~Mei, and J.~Luo.
\newblock Learning multi-attention convolutional neural network for
  fine-grained image recognition.
\newblock In {\em ICCV}, 2017.

\bibitem{bolei}
B.~Zhou, A.~Khosla, A.~Lapedriza, A.~Oliva, and A.~Torralba.
\newblock Learning deep features for discriminative localization.
\newblock In {\em CVPR}, 2016.

\bibitem{fzhou1}
F.~Zhou and Y.~Lin.
\newblock Fine-grained image classification by exploring bipartite-graph
  labels.
\newblock In {\em CVPR}, 2016.

\end{thebibliography}
}

\end{document}